\pdfoutput=1
\documentclass[conference]{ieeeconf}  
\IEEEoverridecommandlockouts                        
\title{\LARGE \bf
Which Physical Tactile Cues Matter for Contact-Rich Imitation Learning? A Controlled Real-Robot Study}

\newcommand{\teaserfig}{%
   \par\vspace{0.1em}
  \noindent\makebox[\textwidth][c]{%
    \begin{minipage}{\textwidth}
      \centering
      \includegraphics[
        width=\linewidth,
        height=6.0cm,
        keepaspectratio
      ]{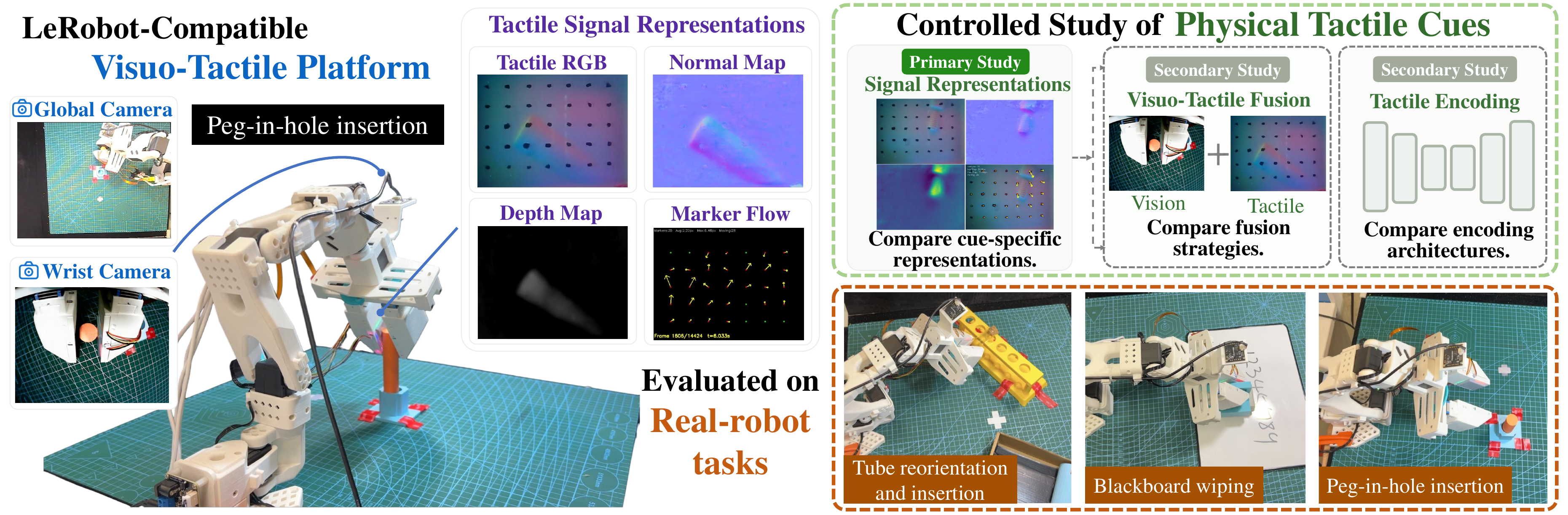}
      \captionof{figure}{This work investigates how tactile signals should be represented, encoded, and fused for contact-rich imitation learning.  We build a LeRobot-compatible visuo-tactile platform that converts the same tactile observation into multiple representations, including tactile RGB, normal/depth maps, and marker-flow fields. Using this platform, we perform controlled comparisons of tactile representations, tactile encoders, and visuo-tactile fusion strategies across three real-robot manipulation tasks.
}
    \label{fig:teaser}
    \end{minipage}%
  }
  \par\vspace{-0.8em}
}

\makeatletter

\def\@IEEEauthorblockconfadjspace{-1.5em}

\let\old@maketitle\@maketitle
\def\@maketitle{%
  \old@maketitle
  \teaserfig
}
\makeatother

\author{
Xuhao Qin,
Xiyan Huang,
Haoyu Dong,
Xi Zhang,
and Chenxi Xiao$^\dagger$
\thanks{$^\dagger$ denotes the corresponding author.}%
\thanks{
This work was supported by the Natural Science Foundation of Shanghai
(Grant No. 25ZR1402370), and partially Key Laboratory of Intelligent Perception and Human-Machine Collaboration (ShanghaiTech University), Ministry of Education.
The experiments were supported by the Core Facility Platform of Computer
Science and Communication (SIST) and GenAI platform, ShanghaiTech University.
}%
\thanks{
Xuhao Qin, Xiyan Huang, Haoyu Dong, Xi Zhang, and Chenxi Xiao are with the
School of Information Science and Technology, ShanghaiTech University,
Shanghai, China.
{\tt\footnotesize \{qinxh2024, origami, donghy2022, zhangxi2024, xiaochx\}@shanghaitech.edu.cn}
}
\thanks{ Project webpage: \href{https://xuhaoqin02.github.io/lerobot-tactile/}{https://xuhaoqin02.github.io/lerobot-tactile/} }
}

\usepackage{adjustbox} 

\usepackage{algorithm}
\usepackage[utf8]{inputenc}
\usepackage{newunicodechar}
\newunicodechar{，}{,}
\usepackage{bbm}
\usepackage{etoolbox}
\usepackage{multicol}
\usepackage{cite}
\usepackage{graphics} %
\usepackage{times} %
\usepackage{amsmath} %
\usepackage{amssymb}  %

\usepackage{svg}
\usepackage{mathrsfs}
\usepackage{amsfonts}
\usepackage{wrapfig}
\usepackage{amsmath}

\usepackage{kantlipsum}
\usepackage{subcaption}
\usepackage{marvosym}

\usepackage{stfloats}
\usepackage{float}
\usepackage{ctable}
\usepackage{cuted}
\usepackage{colortbl}
\usepackage{multirow}
\usepackage{wrapfig}
\usepackage[misc,geometry]{ifsym}
\usepackage{algorithm}
\usepackage{algpseudocode}
\usepackage{xcolor}   %
\usepackage{pifont}    %
\usepackage{graphicx}
\usepackage{caption}
\usepackage{subcaption}

\usepackage{arydshln}
\usepackage{array}
\usepackage{bm}
\usepackage{caption}
\usepackage{setspace}
\let\labelindent\relax
\usepackage{enumitem}
\usepackage{threeparttable}
\usepackage{makecell}  
\usepackage{booktabs,array}
\usepackage{float} 
\usepackage{cuted}
 \usepackage{capt-of}
\usepackage{booktabs}
\usepackage{multirow}
\usepackage{graphicx}

\newcolumntype{L}[1]{>{\centering\arraybackslash}m{#1}}

\newcommand{\TODO}[1][]{\textcolor{red}{\bf [TODO]}}
\setlist[enumerate,1]{itemsep=3pt}

\definecolor{formalgreen}{rgb}{0.1, 0.7, 0.1}  %
\definecolor{formalred}{rgb}{0.9, 0.2, 0.2}  %

\makeatletter
\let\NAT@parse\undefined
\makeatother
\usepackage{hyperref} %
\hypersetup{
  colorlinks   = true,    %
  linkcolor    = red,    %
  citecolor    = blue,      %
}

\begin{document}

\maketitle

 \setcounter{figure}{1}

\thispagestyle{empty}
\pagestyle{empty}

\begin{abstract}
Vision-based tactile sensors capture multiple aspects of physical interaction within a single image stream, including contact appearance, geometry, indentation depth, and tangential deformation. However, existing visuo-tactile policies often directly process tactile observations as images, making it difficult to determine which physical cues contribute to policy learning. This paper presents a controlled real-robot study that compares physically interpretable tactile signals using the LeRobot platform, enhanced with our visuo-tactile perception system. From the same tactile stream, we derive tactile RGB images, normal/depth maps, and marker-flow fields, representing contact appearance, geometry, and tangential deformation, respectively. Using identical task demonstrations and the same policy backbone, we evaluate the individual and combined contributions of these tactile cues. We further compare tactile encoders and visuo-tactile fusion strategies. 
Evaluation across 720 real-robot trials yields four key findings: 
(1) tactile perception substantially improves the overall success rate from 57.8\% to 78.9\%, although its contribution varies across task stages; 
(2) marker flow is the most effective individual cue, achieving a success rate of 74.4\% and playing a  particularly important role in regulating established contact;  
(3) cue-factorized encoding consistently improves performance, highlighting that different physical cues require tailored processing techniques; and
(4) although the effectiveness of tactile fusion is task-dependent, cross-attention offers a lightweight solution with performance comparable to that of pretrained tactile encoders.

\end{abstract}

\vspace{-1mm}
\section{Introduction}

Visuomotor imitation learning has achieved strong performance across a wide range of robotic manipulation tasks~\cite{chi2025diffusion}. However, successful manipulation often depends on local physical interaction states that are only partially observable by external cameras. Contact regions may be occluded by the robot or the manipulated object, while subtle indentation, tangential deformation, slip, and contact stability may remain visually ambiguous. Tactile feedback helps bridge this gap by directly observing local interactions and has therefore become increasingly important for contact-rich manipulation~\cite{calandra2018more,dong2021tactile,yu2024mimictouch,xue2025reactive}.

In this paper, we focus on vision-based tactile sensors, which image the deformation of a compliant elastomer to provide dense observations of the contact interface~\cite{yuan2017gelsight,lambeta2020digit,taylor2022gelslim}. Unlike conventional visual images, tactile images arise from physical interaction and encode several distinct forms of contact information. RGB images capture tactile appearance, including contact patterns and color variations induced by elastomer deformation; surface-normal and depth maps describe contact geometry; and marker-displacement fields capture tangential deformation. We refer to these explicit input forms as \textbf{\emph{tactile cue representations}}, distinguishing them from the latent representations learned by tactile encoders. Despite their different physical meanings, tactile observations are commonly processed as raw RGB images using generic image encoders, leaving unclear which underlying signals support policy learning.

Prior work has investigated tactile-specific encoders, visuo-tactile fusion, and broader policy-learning benchmarks~\cite{higuera2024sparsh, gupta2025sensorinvariant,heng2025vitacformerlearningcrossmodalrepresentation,luu2025manifeel}. These studies primarily examine how complete tactile observations are encoded, integrated, or evaluated, rather than isolating the contributions of their constituent physical signals. In contrast, we systematically evaluate the individual and joint roles of tactile appearance, contact geometry, and tangential deformation derived from the same tactile observations under matched learning conditions. Specifically, we ask which signals are most informative for policy learning and whether their combinations provide complementary benefits. Beyond informing representation choice, this analysis offers practical guidance for tactile-sensor design by identifying which tactile cues should be prioritized under different contact regimes and when multiple cues should be jointly retained.

To address these questions, we derive RGB, surface-normal/depth, marker-flow, and combined signal representations from the same real-world tactile stream. We compare these representations using identical demonstrations, visual observations, and policy backbones. We then fix the tactile input to the combined representation and compare generic image encoders with tactile-pretrained encoders, as well as concatenation-based and cross-attention-based fusion. This staged experimental design compares tactile input representations and evaluates encoder and fusion choices under fixed full-tactile input.

Our experiments yield these main findings. First, tactile perception substantially improves overall policy performance, although the utility of individual tactile cues varies across interaction stages.
Second, marker flow is the strongest individual cue overall, with qualitative failure analysis suggesting its potential role in regulating established contact, whereas normal/depth cues appear more useful for contact establishment. Although full tactile provides only a modest improvement over marker flow in final success rate, combining these cues promotes smoother and more continuous transitions across interaction stages. Third, while the effectiveness of visuo-tactile fusion remains task-dependent, cue-factorized encoding, and cross-attention consistently improve performance, and provide a lightweight alternative with performance comparable to tactile-pretrained encoding.

In summary, this work makes three contributions:
\begin{enumerate}
    \setlength{\itemsep}{0pt}
    \setlength{\parskip}{0pt}
    \setlength{\parsep}{0pt}

    \item We introduce an open, LeRobot-compatible visuo-tactile testbed with a standardized Diffusion Policy pipeline for
    reproducible evaluation of tactile signal representations.
    
    \item We conduct a comparative study of the individual and joint roles of tactile appearance, contact geometry, and tangential deformation, showing that marker flow is the strongest individual cue overall and that the utility of different tactile cues varies across interaction stages.

    \item We systematically compare generic and tactile-pretrained encoders and visuo-tactile fusion strategies with fixed tactile inputs, showing that cross-attention offers a simple, lightweight option that balances performance and complexity.
\end{enumerate}

\begin{figure*}[!t]
    \centering
    \includegraphics[width=\linewidth]{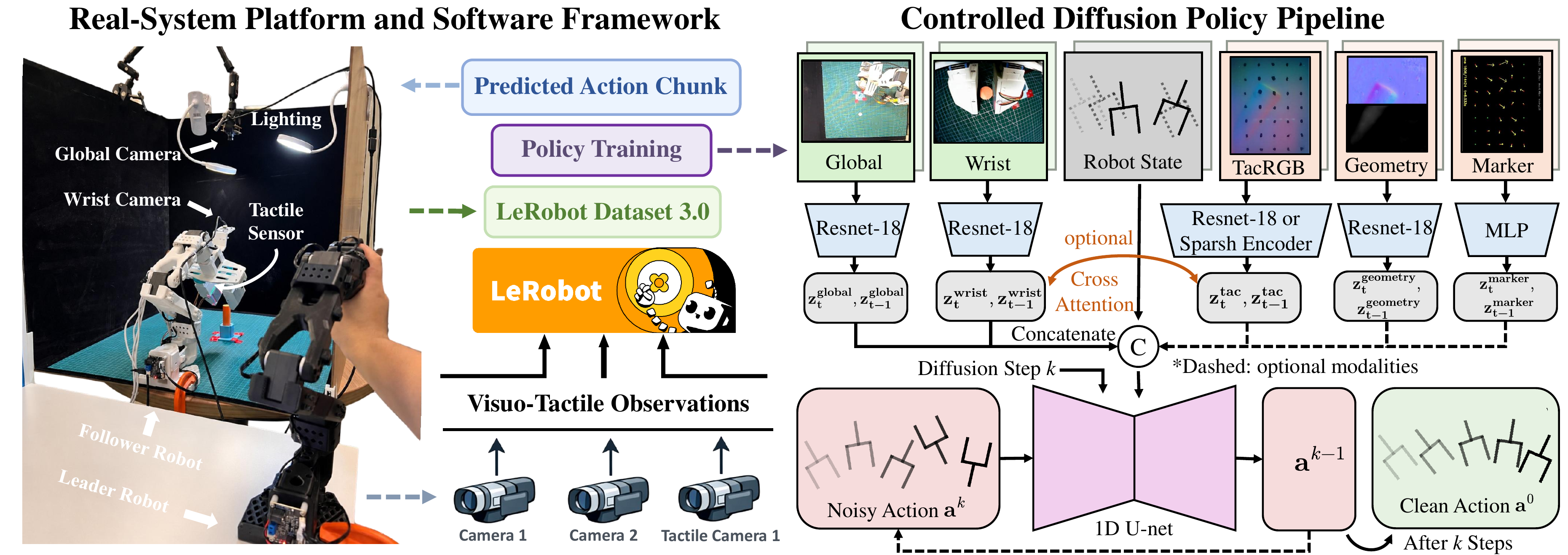}
    \caption{Overview of the LeRobot-compatible visuo-tactile system and Diffusion Policy pipeline. Tactile RGB and geometry use image encoders, while marker flow uses an MLP. Wrist-camera and tactile-RGB features can optionally interact through cross-attention before multimodal fusion and action prediction. Only the tactile-RGB encoder is varied in the encoding study.}
    \label{fig:system_overview}
    \vspace{-0.5cm}
\end{figure*}

\section{Related Work}

\subsection{Vision-based Tactile Sensors}

Vision-based tactile sensors have become an important sensing modality for robotic manipulation. Representative sensors, including GelSight~\cite{yuan2017gelsight}, GelSlim~\cite{taylor2022gelslim}, and DIGIT~\cite{lambeta2020digit}, provide dense, localized contact information by imaging the deformation of a compliant surface. Existing sensors adopt different sensing recipes that expose distinct physical cues: photometric designs use controlled illumination and reflective elastomers to reconstruct contact geometry through photometric stereo~\cite{johnson2009retrographic}, whereas marker-based sensors such as FingerVision track dot motion to capture force- and slip-related cues~\cite{yamaguchi2019tactile}, and TacForce maps marker deformation to dense force distributions~\cite{duong2024tacforce}. Marker-free designs emphasize tactile RGB appearance, while hybrid designs combine photometric geometry and marker motion. However, which combination of appearance, geometry, and deformation cues best supports policy learning remains unclear. Our sensor combines calibrated illumination with a marker-embedded elastomer, allowing these cues to be derived from the same tactile stream and compared under matched conditions.

\vspace{-2mm}
\subsection{Visuo-Tactile Learning for Contact-Rich Manipulation}
\vspace{-0.5mm}
Existing visuo-tactile manipulation methods differ in when and how tactile information is incorporated into the policy and control loop. Early work demonstrated that tactile sensing improves grasping, regrasping, and insertion by providing contact information unavailable from vision alone~\cite{calandra2018more,dong2021tactile}. More recent imitation-learning systems have leveraged tactile feedback for fine-grained assembly, reactive policy learning, and contact-aware manipulation. MimicTouch~\cite{yu2024mimictouch} uses multimodal human demonstrations with tactile feedback for contact-rich tasks, while Reactive Diffusion Policy~\cite{xue2025reactive} introduces a slow--fast visuo-tactile architecture for rapid tactile responses. Although these studies demonstrate different ways of exploiting tactile feedback, they typically rely on a fixed tactile representation tailored to a particular sensor, task, or policy architecture. Building on their evidence that touch benefits contact-rich manipulation, we investigate which tactile cues are most useful for imitation learning.

\vspace{-2mm}
\subsection{Tactile Representation Learning and Visuo-Tactile Fusion}
\vspace{-0.5mm}
Recent work has moved beyond handcrafted tactile features toward tactile-specific representation learning. Sparsh~\cite{higuera2024sparsh} learns general-purpose tactile representations through self-supervised pretraining on vision-based tactile data, while SITR~\cite{gupta2025sensorinvariant} investigates sensor-invariant representations across sensors and tasks. In parallel, visuo-tactile fusion models study how local tactile cues should interact with visual scene information. ViTacFormer~\cite{heng2025vitacformerlearningcrossmodalrepresentation} uses cross-modal attention to learn coupled visual and tactile representations, ManiFeel~\cite{luu2025manifeel} benchmarks tactile modalities, encoders, and policy architectures across simulated and real-world tasks, and CONTACT~\cite{saka2026contact} compares tactile RGB with an aggregated normal--shear force-field representation for robotic disassembly. Complementing these studies, we compare tactile RGB, normal/depth geometry, marker-flow, and their combination under identical learning conditions; and subsequently, examine the performance of different design choices for visuo-tactile fusion.

\vspace{-2mm}
\section{Method}
\vspace{-1mm}
\subsection{System Overview}
\vspace{-2mm}

The overall framework is shown in Fig.~\ref{fig:system_overview}, which consists of a LeRobot platform augmented with visuo-tactile sensing for data collection, and a Diffusion Policy pipeline. The platform synchronously records global-camera images, wrist-camera images, robot proprioception, actions, and tactile observations. The same tactile stream is then converted into tactile RGB, normal/depth, and marker-flow representations. Within the shared policy framework, these representations are processed by modality-specific frontends and fused with visual and proprioceptive features for action prediction. This unified design enables controlled comparisons of tactile cue representations, tactile-RGB encoders, and visuo-tactile fusion strategies under matched demonstrations and policy settings,while modality-specific frontends differ across some tactile input configurations.

\vspace{-3mm}

\subsection{Hardware Platform and Data Collection}
\vspace{-1mm}
As illustrated in Fig.~\ref{fig:system_overview}, we build a LeRobot-compatible visuo-tactile parallel-jaw gripper for collecting contact-rich demonstrations. One fingertip is equipped with a vision-based tactile sensor consisting of a marker-embedded elastomer, an internal camera, and calibrated illumination, while the opposite jaw provides passive mechanical support. The tactile sensor is integrated into the gripper structure to capture contact-induced deformation during manipulation.

During demonstration collection, the system synchronizes global-camera images, wrist-camera images, robot proprioception, actions, and tactile observations from the tactile-equipped fingertip within the LeRobot framework. These multimodal observations are stored in a unified LeRobot dataset format. The same tactile stream is subsequently converted into multiple physically interpretable representations for the controlled studies described below.

\subsection{Study of Physical Tactile Cues}
\begin{figure}[t]
    \centering
    \includegraphics[width=\linewidth]{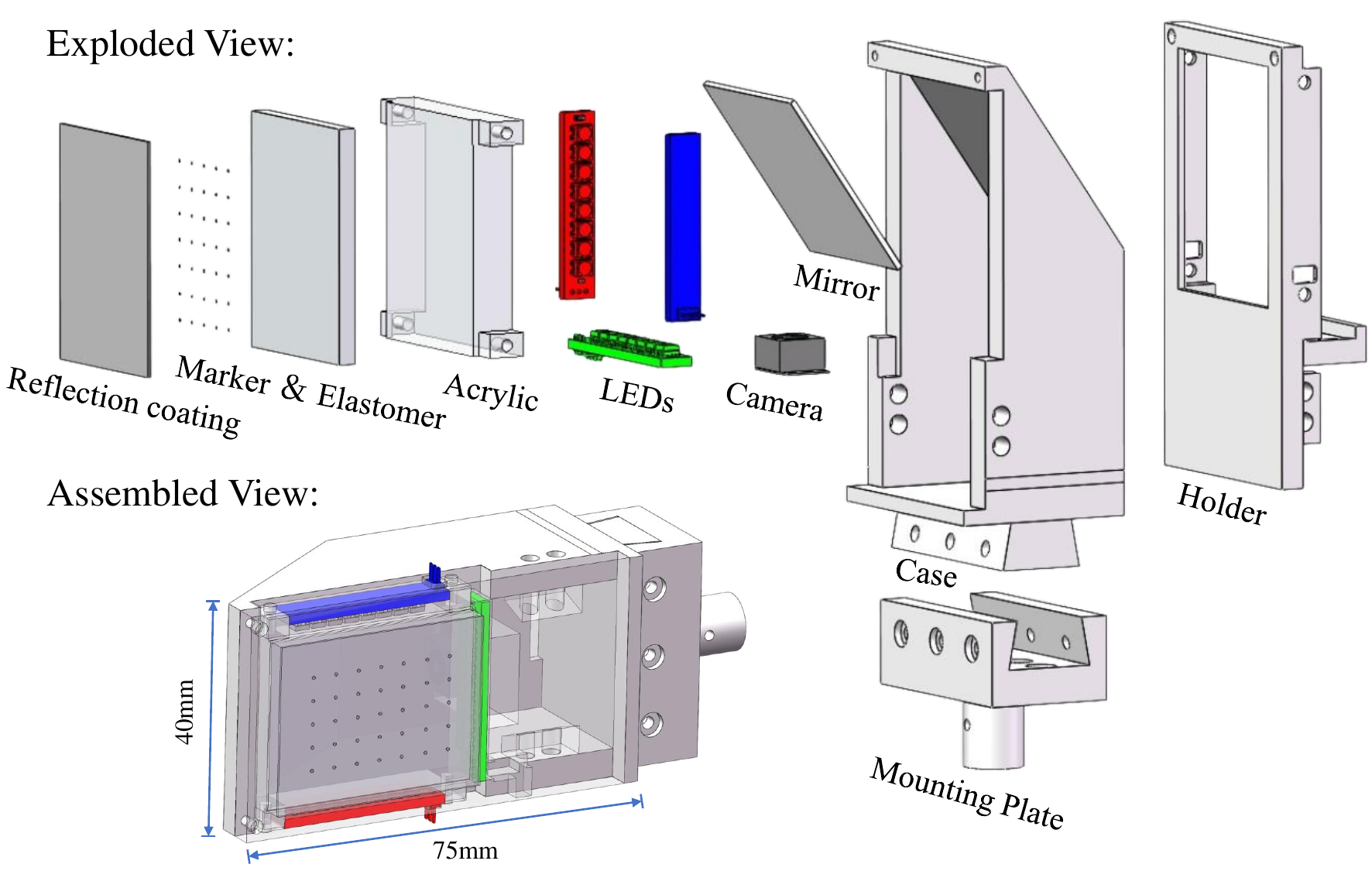}
    \caption{Sensor design, including an exploded view of the internal components, and the assembled CAD model.}
    \label{fig:tactile_gripper}
    \vspace{-0.5cm}
\end{figure}

\begin{figure}[t]
    \centering
    \includegraphics[width=\linewidth]{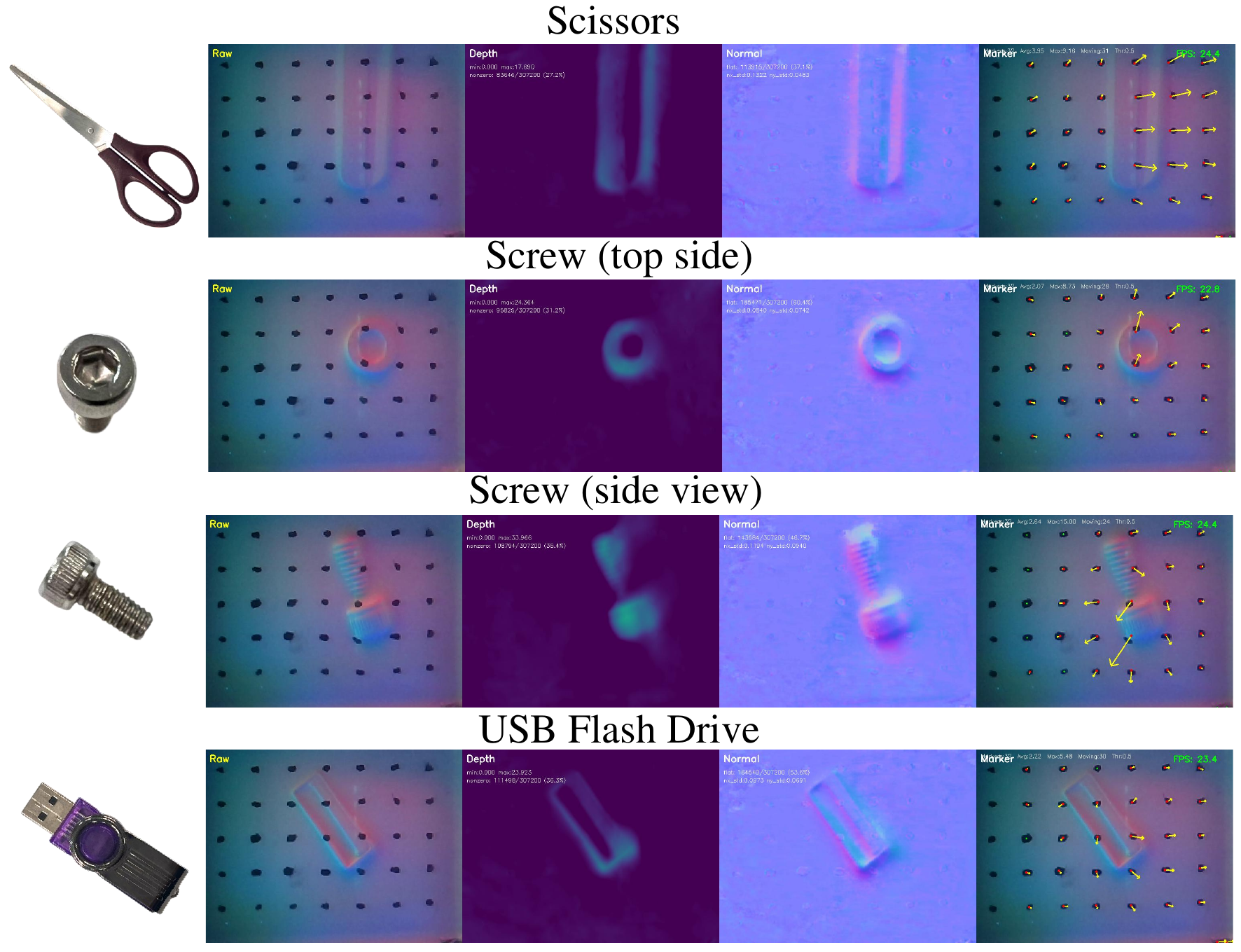}
    \caption{Examples of tactile representations for different contact geometries. From left to right, each row shows the contacted object, tactile RGB image, depth map, surface-normal map, and marker-flow field.}
    \label{fig:tactile_representation_examples}
    \vspace{-0.5cm}
\end{figure}

\vspace{-2mm}
Our first study examines how different physical tactile cues contribute to contact-rich policy learning. Following the sensing principle of GelSight Wedge~\cite{wang2021gelsight}, we developed a sensor that combines a marker-embedded elastomer, an internal camera, and calibrated illumination (Fig.~\ref{fig:tactile_gripper}). From the same tactile stream, we derive three physically interpretable cue representations, as shown in Fig.~\ref{fig:tactile_representation_examples}: tactile appearance, contact geometry, and tangential deformation.

\indent\textbf{Tactile Appearance:}\quad
Let \(I_t \in \mathbb{R}^{H \times W \times 3}\) denote the raw tactile RGB image at time \(t\), and let \(I_0\) denote the no-contact reference image. We use \(I_t\) directly as the tactile-appearance input, preserving contact patterns and color variations induced by local elastomer deformation (including markers).

\indent\textbf{Contact Geometry:}\quad
To recover contact geometry, we first compute the normalized reference-subtracted tactile image \(\tilde{I}_t=\mathcal{N}(I_t-I_0)\). A neural calibration network then estimates a dense surface-normal map \(N_t=f_{\theta}(\tilde{I}_t)\), where \(N_t\in\mathbb{R}^{H\times W\times 3}\) and \(f_{\theta}\) is trained using spherical indentation data with analytically computed surface normals. The predicted normal field is further integrated through Poisson reconstruction to obtain a depth map \(D_t\in\mathbb{R}^{H\times W\times 1}\), which describes local indentation of the elastomer surface.

\indent\textbf{Tangential Deformation:}\quad
To capture contact-induced tangential deformation, we track embedded markers on the tactile surface. For marker \(i\), its image-plane displacement is defined as \(\Delta m_i^t=m_i^t-m_i^0\), where \(m_i^0,m_i^t\in\mathbb{R}^{2}\) denote its reference and current positions, respectively. The sparse marker displacements are interpolated into a dense marker-flow field \(F_t\in\mathbb{R}^{H\times W\times 2}\), which describes tangential deformation of the tactile surface.

\indent\textbf{Tactile Input Configurations:}\quad
We evaluate four tactile input configurations:
\vspace{-1mm}
\begin{equation}
o_t^\tau =
\begin{cases}
I_t, & \text{tactile RGB},\\
[N_t,D_t], & \text{normal/depth},\\
F_t, & \text{marker flow},\\
\{I_t,[N_t,D_t],F_t\}, & \text{full tactile}.
\end{cases}
\end{equation}

Here, $[\cdot,\cdot]$ denotes channel-wise concatenation within a single representation, whereas $\{\cdot,\cdot\}$ denotes a collection of modality inputs that are encoded by separate frontends and subsequently fused by the policy.
The full-tactile configuration combines tactile appearance, contact geometry, and tangential deformation derived from the same sensor stream.

\vspace{-1mm}
\subsection{Studies of Tactile Encoding and Visuo-Tactile Fusion}
\vspace{-1mm}
We use Diffusion Policy~\cite{chi2025diffusion} as the shared imitation-learning backbone. At each time step, the visual observation is \(o_t^v=\{I_t^g,I_t^w\}\), where \(I_t^g\) and \(I_t^w\) denote the global- and wrist-camera images, respectively. The full observation is \(o_t=\{o_t^v,o_t^\tau,q_t\}\), comprising vision, one tactile configuration, and robot proprioception \(q_t\).

 Given an observation history of length 4, the policy predicts 24 future actions and executes the first 8 actions before replanning. All studies use the same observations, action space, policy backbone, horizons, training data, and training budget.

 With the full-tactile input fixed to \(\{I_t,[N_t,D_t],F_t\}\), we first compare a generic ResNet-18 \cite{he2016deep}encoder with the tactile-pretrained Sparsh-DINO encoder~\cite{higuera2024sparsh} for the tactile RGB branch, while keeping the normal/depth and marker-flow frontends unchanged. We then compare feature concatenation with cross-attention fusion~\cite{heng2025vitacformerlearningcrossmodalrepresentation}. Concatenation aggregates independently encoded features, whereas cross-attention explicitly models interactions between wrist-camera and tactile-RGB features before multimodal fusion. These comparisons isolate the effects of tactile encoding and visuo-tactile fusion under matched conditions.

\begin{figure*}[t]
    \centering
    \includegraphics[width=\linewidth]{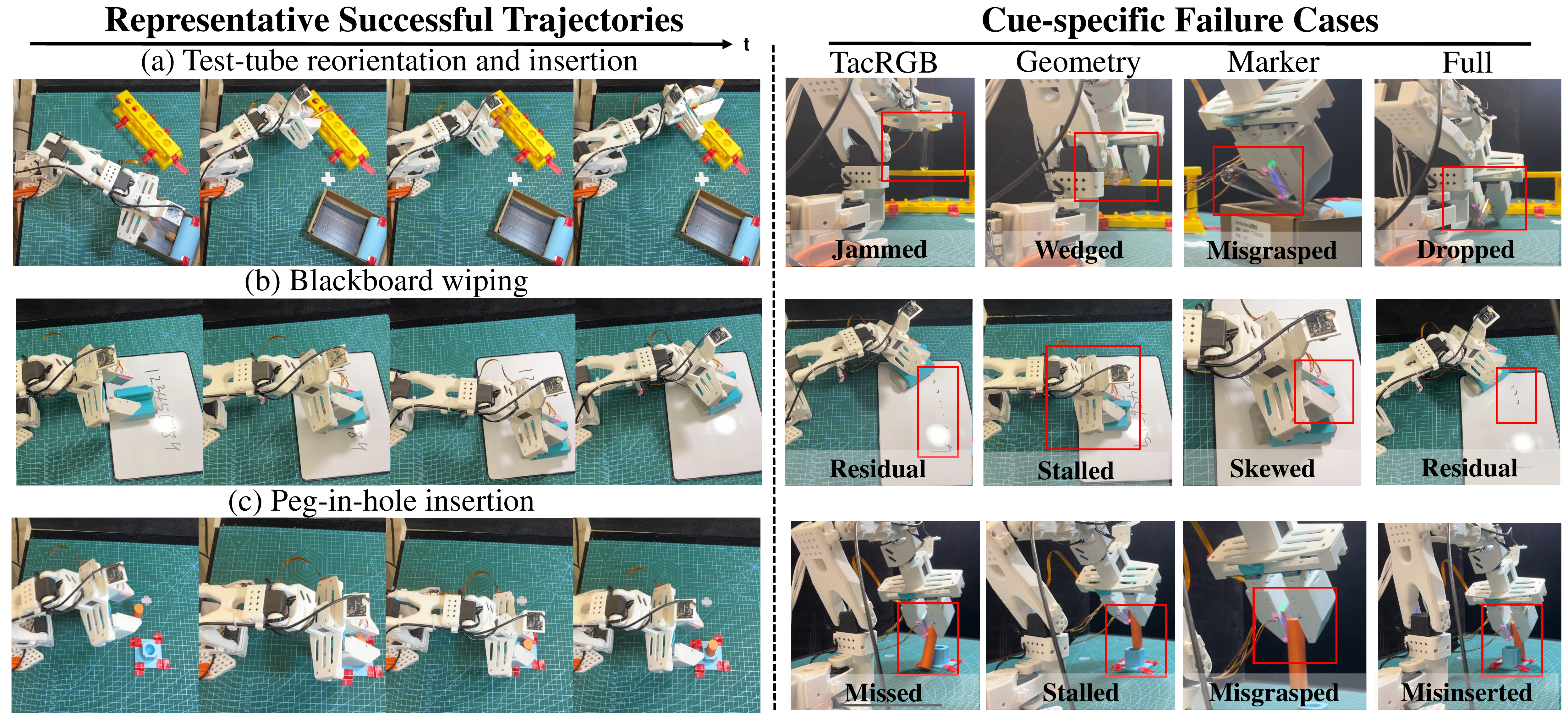}
    \vspace{-0.5cm} 
    \caption{Representative successful trajectories and cue-specific failure cases for three real-robot contact-rich manipulation tasks. The left panel shows successful executions over time, while the right panel presents representative failures under tactile RGB, normal/depth geometry, marker-flow, and full-tactile inputs, illustrating how different tactile cues are associated with failures at different interaction stages.}
    \label{fig:process_analysis}

    \vspace{-0.3cm}
\end{figure*}

\section{Experiments}
\label{sec:experiments}

We evaluate the effect of different tactile signal representations on three real-robot contact-rich manipulation tasks: test-tube reorientation and insertion, blackboard wiping, and peg-in-hole insertion. 
These tasks involve different contact regimes, including in-gripper object adjustment, sustained surface contact, tangential motion, and tight-clearance insertion.
The experiments are designed to answer three questions:
\textbf{Q1:} Does tactile feedback improve performance beyond dual-camera vision alone?
\textbf{Q2:} Which physical tactile cues provide useful information for contact-rich imitation learning?
\textbf{Q3:} In the full-tactile setting, how do tactile encoding and visuo-tactile fusion strategies affect performance?

\vspace{-1mm}

\subsection{Experimental Setup}
\label{sec:experimental_setup}

All demonstrations are collected using our LeRobot-compatible visuo-tactile pipeline and contain synchronized global-camera
images, wrist-camera images, robot proprioception, actions, and tactile observations. The same tactile stream is converted into tactile RGB, normal/depth, marker-flow, and full-tactile inputs, and all policy variants for each task use the same demonstration set.

The three tasks are designed to emphasize different forms of contact reasoning. In the test-tube reorientation and insertion task, the robot must adjust the pose of a tube inside the gripper and insert it into a holder. This task requires the policy to infer local in-gripper contact changes that may not be fully visible from external cameras. In the blackboard wiping task, the robot must maintain stable contact with a surface while moving tangentially across the target region, making pressure regulation and shear-related cues important. In the peg-in-hole insertion task, the robot must align and insert a peg under tight clearance, where small contact errors can lead to jamming or failure.

\begin{table*}[t]
    \centering
    \setlength{\tabcolsep}{6pt}
    \renewcommand{\arraystretch}{1.1}

    \caption{Controlled comparison of tactile inputs under matched visual observations, policy backbone, fusion, demonstrations, and training settings. Values are success rates over 30 real-robot trials per task. Total success rates are aggregated over 90 trials across the three tasks.}
    \label{tab:tactile_representation}
    \resizebox{0.92\textwidth}{!}{
    \begin{tabular}{lcccc}
        \toprule
        \textbf{Method}
        & \textbf{Tube Reorientation}
        & \textbf{Blackboard Wiping}
        & \textbf{Peg-in-hole Insertion}
        & \textbf{Total Success Rate} \\
        \midrule
        Vision only
        & 14/30 (46.7\%)
        & 18/30 (60.0\%)
        & 20/30 (66.7\%)
        & 52/90 (57.8\%) \\

        Vision + tactile RGB
        & 22/30 (73.3\%)
        & 24/30 (80.0\%)
        & 20/30 (66.7\%)
        & 66/90 (73.3\%) \\

        Vision + normal/depth
        & 20/30 (66.7\%)
        & 19/30 (63.3\%)
        & 21/30 (70.0\%)
        & 60/90 (66.7\%) \\

        Vision + marker flow
        & 22/30 (73.3\%)
        & 23/30 (76.7\%)
        & 22/30 (73.3\%)
        & 67/90 (74.4\%) \\

        Vision + full tactile
        & $\mathbf{23/30 (76.7\%)}$
        & $\mathbf{25/30 (83.3\%)}$
        & $\mathbf{23/30 (76.7\%)}$
        & $\mathbf{71/90 (78.9\%)}$ \\
        \bottomrule
    \end{tabular}
    }
    \vspace{-0.7cm}
\end{table*}

\vspace{-1mm}
\subsection{Controlled Evaluation Protocol}
\label{sec:evaluation_protocol}

We conduct two controlled evaluations. The first compares tactile RGB, normal/depth, marker flow, and full tactile representations while keeping all other components and settings fixed. ResNet-18 encodes tactile RGB and normal/depth, while an MLP processes marker flow. A vision-only policy serves as the baseline. Tactile RGB denotes raw images containing markers. Unlike tactile RGB, the full tactile representation factorizes and separately encodes each cue using its corresponding network.

 \vspace{-1mm}
The second evaluation fixes the full-tactile input and compares two tactile-RGB encoders, ResNet-18 and Sparsh-DINO, together with feature concatenation and cross-attention fusion. The normal/depth and marker-flow frontends remain unchanged.
\vspace{-1mm}
For each policy configuration and task, we conduct 30 real-robot trials, resulting in 720 trials across the eight distinct policy configurations reported in Tables~\ref{tab:tactile_representation} and~\ref{tab:encoder_fusion}. A test-tube trial succeeds when the tube is reoriented and inserted into the holder. A wiping trial succeeds when the target region is wiped while stable surface contact is maintained. A peg-in-hole trial succeeds when the peg is inserted without jamming or dropping.

\vspace{-1.9mm}
\subsection{Results and Analysis}
\label{sec:results}

Table~\ref{tab:tactile_representation} shows that tactile feedback generally improves performance over vision alone. Across the three tasks, full tactile achieves 71/90 successes (78.9\% $\pm$ 8.4\%), compared with 52/90 (57.8\% $\pm$ 10.2\%) for vision only, yielding an aggregate gain of 21.1 percentage points. The task-level gains are 30.0 points for test-tube manipulation, 23.3 points for wiping, and 10.0 points for peg insertion. Among individual cues, marker flow achieves the highest aggregate performance at 67/90 (74.4\% $\pm$ 9.0\%), although cue effectiveness remains task dependent: tactile RGB and marker flow tie on the test-tube task, tactile RGB performs best on wiping, and marker flow performs best on peg insertion. Full tactile achieves the highest success rate on all three tasks, but exceeds the strongest individual cue by only one success per task. Although the gain in final success rate is modest, combining cues improves continuity across contact establishment, regulation, and task completion. Thus, full tactile consistently outperforms the baselines, demonstrating the benefits of combining complementary cues and factorized encoding.

\vspace{-0.5mm}
Table~\ref{tab:encoder_fusion} further evaluates encoding and fusion under the fixed full-tactile input. With ResNet-18, cross-attention raises aggregate success from 71/90 (78.9\% $\pm$ 8.4\%) to 77/90 (85.6\% $\pm$ 7.3\%), a gain of 6.7 percentage points. With Sparsh-DINO, it improves success from 75/90 (83.3\% $\pm$ 7.7\%) to 79/90 (87.8\% $\pm$ 6.8\%), a gain of 4.4 points. Cross-attention therefore provides numerical gains across both encoders. Notably, ResNet-18 with cross-attention achieves performance comparable to Sparsh-DINO with concatenation. Encoder effects are less consistent: Sparsh-DINO improves test-tube manipulation under both fusion strategies, produces no gain on wiping, and performs slightly below ResNet-18 with cross-attention on peg insertion. These results suggest that cross-attention provides a lightweight alternative when tactile-specific pretraining is unavailable.

Qualitative inspection of the dominant failure modes, illustrated by representative cases in Fig.~\ref{fig:process_analysis}, suggests a stage-wise distinction among tactile cues. In test-tube manipulation and wiping, normal/depth policies often establish a viable grasp or surface contact but fail during subsequent rotation or sustained motion. Marker-flow policies show the complementary tendency: initial grasping is less reliable, but established contact is more effectively regulated through tangential interaction. Peg insertion is less tolerant to these initial grasp errors, leading marker-flow policies to produce more off-center or unstable grasps. Tactile RGB exhibits more distributed failures across stages.  

\vspace{-2mm}
\section{Conclusion}
\label{sec:conclusion}

This paper presents a comparative real-robot study of physical tactile cues for contact-rich imitation learning. Across 720 real-robot trials, incorporating tactile perception increases the aggregate success rate from 57.8\% $\pm$ 10.2\% to 78.9\% $\pm$ 8.4\%, although the contribution of touch varies across interaction stages. Among the evaluated cues, marker flow is the strongest individual cue overall, achieving a 74.4\% $\pm$ 9.0\% success rate and playing an important role in force regulation after contact is established. Full tactile input outperforms marker flow by only 4.4 percentage points, with its primary benefit being smoother, more continuous transitions between task-execution stages. Cue-factorized encoding also outperforms direct tactile-RGB encoding.  Although the effectiveness of visuo-tactile fusion is task-dependent, cross-attention offers a lightweight alternative with performance comparable to that of tactile-pretrained encoding. However, the study is limited by the absence of direct force measurements, differences in frontend architectures, and evaluation on a single tactile sensor, a parallel-jaw gripper, and three tabletop tasks. Future work will investigate the physical roles of tactile representations and their generalization across sensors, objects, and robot embodiments.

\begin{table*}[t]
    \centering
    \setlength{\tabcolsep}{5pt}
    \renewcommand{\arraystretch}{1.15}

    \caption{
    Controlled comparison of tactile encoder choices and visuo-tactile fusion strategies in the full-tactile setting. All methods use tactile RGB, normal/depth, and marker-flow inputs. Task-level success rates are reported over 30 trials, and total success rates are aggregated over 90 trials across the three tasks.
    }
    \label{tab:encoder_fusion}
    \resizebox{\textwidth}{!}{
    \begin{tabular}{llcccc}
        \toprule
        \textbf{Tactile Encoder}
        & \textbf{Fusion Strategy}
        & \textbf{Tube Reorientation}
        & \textbf{Blackboard Wiping}
        & \textbf{Peg-in-hole Insertion}
        & \textbf{Total Success Rate} \\
        \midrule
        ResNet
        & Concat.
        & 23/30 (76.7\%)
        & 25/30 (83.3\%)
        & 23/30 (76.7\%)
        & 71/90 (78.9\%) \\

        ResNet
        & Cross-Attn.
        & 24/30 (80.0\%)
        & 27/30 (90.0\%)
        & $\mathbf{26/30 (86.7\%)}$
        & 77/90 (85.6\%) \\

        Sparsh-DINO
        & Concat.
        & 26/30 (86.7\%)
        & 25/30 (83.3\%)
        & 24/30 (80.0\%)
        & 75/90 (83.3\%) \\

        Sparsh-DINO
        & Cross-Attn.
        & $\mathbf{27/30 (90.0\%)}$
        & $\mathbf{27/30 (90.0\%)}$
        & 25/30 (83.3\%)
        & $\mathbf{79/90 (87.8\%)}$ \\
        \bottomrule
    \end{tabular}
    }
    \vspace{-0.3cm}
\end{table*}

{\small
\bibliographystyle{IEEEtran}
\bibliography{reference}

\begin{thebibliography}{10}
\providecommand{\url}[1]{#1}
\csname url@samestyle\endcsname
\providecommand{\newblock}{\relax}
\providecommand{\bibinfo}[2]{#2}
\providecommand{\BIBentrySTDinterwordspacing}{\spaceskip=0pt\relax}
\providecommand{\BIBentryALTinterwordstretchfactor}{4}
\providecommand{\BIBentryALTinterwordspacing}{\spaceskip=\fontdimen2\font plus
\BIBentryALTinterwordstretchfactor\fontdimen3\font minus
  \fontdimen4\font\relax}
\providecommand{\BIBforeignlanguage}[2]{{%
\expandafter\ifx\csname l@#1\endcsname\relax
\typeout{** WARNING: IEEEtran.bst: No hyphenation pattern has been}%
\typeout{** loaded for the language `#1'. Using the pattern for}%
\typeout{** the default language instead.}%
\else
\language=\csname l@#1\endcsname
\fi
#2}}
\providecommand{\BIBdecl}{\relax}
\BIBdecl

\bibitem{chi2025diffusion}
C.~Chi, Z.~Xu, S.~Feng, E.~Cousineau, Y.~Du, B.~Burchfiel, R.~Tedrake, and
  S.~Song, ``Diffusion policy: Visuomotor policy learning via action
  diffusion,'' \emph{The International Journal of Robotics Research}, vol.~44,
  no. 10-11, pp. 1684--1704, 2025.

\bibitem{calandra2018more}
R.~Calandra, A.~Owens, D.~Jayaraman, J.~Lin, W.~Yuan, J.~Malik, E.~H. Adelson,
  and S.~Levine, ``More than a feeling: Learning to grasp and regrasp using
  vision and touch,'' \emph{IEEE Robotics and Automation Letters}, vol.~3,
  no.~4, pp. 3300--3307, 2018.

\bibitem{dong2021tactile}
S.~Dong, D.~K. Jha, D.~Romeres, S.~Kim, D.~Nikovski, and A.~Rodriguez,
  ``Tactile-rl for insertion: Generalization to objects of unknown geometry,''
  in \emph{2021 IEEE International Conference on Robotics and Automation
  (ICRA)}.\hskip 1em plus 0.5em minus 0.4em\relax IEEE, 2021, pp. 6437--6443.

\bibitem{yu2024mimictouch}
\BIBentryALTinterwordspacing
K.~Yu, Y.~Han, Q.~Wang, V.~Saxena, D.~Xu, and Y.~Zhao, ``Mimictouch: Leveraging
  multi-modal human tactile demonstrations for contact-rich manipulation,'' in
  \emph{8th Annual Conference on Robot Learning}, 2024. [Online]. Available:
  \url{https://openreview.net/forum?id=7yMZAUkXa4}
\BIBentrySTDinterwordspacing

\bibitem{xue2025reactive}
H.~Xue, J.~Ren, W.~Chen, G.~Zhang, Y.~Fang, G.~Gu, H.~Xu, and C.~Lu, ``Reactive
  diffusion policy: Slow-fast visual-tactile policy learning for contact-rich
  manipulation,'' in \emph{Proceedings of Robotics: Science and Systems (RSS)},
  2025.

\bibitem{yuan2017gelsight}
W.~Yuan, S.~Dong, and E.~H. Adelson, ``Gelsight: High-resolution robot tactile
  sensors for estimating geometry and force,'' \emph{Sensors}, vol.~17, no.~12,
  p. 2762, 2017.

\bibitem{lambeta2020digit}
M.~Lambeta, P.-W. Chou, S.~Tian, B.~Yang, B.~Maloon, V.~R. Most, D.~Stroud,
  R.~Santos, A.~Byagowi, G.~Kammerer \emph{et~al.}, ``Digit: A novel design for
  a low-cost compact high-resolution tactile sensor with application to in-hand
  manipulation,'' \emph{IEEE Robotics and Automation Letters}, vol.~5, no.~3,
  pp. 3838--3845, 2020.

\bibitem{taylor2022gelslim}
I.~H. Taylor, S.~Dong, and A.~Rodriguez, ``Gelslim 3.0: High-resolution
  measurement of shape, force and slip in a compact tactile-sensing finger,''
  in \emph{2022 International Conference on Robotics and Automation
  (ICRA)}.\hskip 1em plus 0.5em minus 0.4em\relax IEEE, 2022, pp.
  10\,781--10\,787.

\bibitem{higuera2024sparsh}
\BIBentryALTinterwordspacing
C.~Higuera, A.~Sharma, C.~K. Bodduluri, T.~Fan, P.~Lancaster, M.~Kalakrishnan,
  M.~Kaess, B.~Boots, M.~Lambeta, T.~Wu, and M.~Mukadam, ``Sparsh:
  Self-supervised touch representations for vision-based tactile sensing,'' in
  \emph{8th Annual Conference on Robot Learning}, 2024. [Online]. Available:
  \url{https://openreview.net/forum?id=xYJn2e1uu8}
\BIBentrySTDinterwordspacing

\bibitem{gupta2025sensorinvariant}
H.~Gupta, Y.~Mo, S.~Jin, and W.~Yuan, ``Sensor-invariant tactile
  representation,'' in \emph{The Thirteenth International Conference on
  Learning Representations}, 2025.

\bibitem{heng2025vitacformerlearningcrossmodalrepresentation}
\BIBentryALTinterwordspacing
L.~Heng, H.~Geng, K.~Zhang, P.~Abbeel, and J.~Malik, ``Vitacformer: Learning
  cross-modal representation for visuo-tactile dexterous manipulation,'' 2025.
  [Online]. Available: \url{https://arxiv.org/abs/2506.15953}
\BIBentrySTDinterwordspacing

\bibitem{luu2025manifeel}
Q.~K. Luu, P.~Zhou, Z.~Xu, Z.~Zhang, Q.~Qiu, and Y.~She, ``Manifeel:
  Benchmarking and understanding visuotactile manipulation policy learning,''
  \emph{arXiv preprint arXiv:2505.18472}, 2025.

\bibitem{johnson2009retrographic}
M.~K. Johnson and E.~H. Adelson, ``Retrographic sensing for the measurement of
  surface texture and shape,'' in \emph{2009 IEEE conference on computer vision
  and pattern recognition}.\hskip 1em plus 0.5em minus 0.4em\relax IEEE, 2009,
  pp. 1070--1077.

\bibitem{yamaguchi2019tactile}
A.~Yamaguchi and C.~G. Atkeson, ``Tactile behaviors with the vision-based
  tactile sensor {FingerVision},'' \emph{International Journal of Humanoid
  Robotics}, vol.~16, no.~3, p. 1940002, 2019.

\bibitem{duong2024tacforce}
L.~Van~Duong and T.~T. Mac, ``{TacForce}: From images to tactile force
  distribution,'' in \emph{2024 IEEE 11th International Conference on
  Computational Cybernetics and Cyber-Medical Systems (ICCC)}.\hskip 1em plus
  0.5em minus 0.4em\relax IEEE, 2024, pp. 317--320.

\bibitem{saka2026contact}
Y.~Saka, J.-C. Hu, A.~Desai, Z.~Zhang, B.~Zhang, Q.~K. Luu, M.~R.~I. Prince,
  M.~Zheng, and Y.~She, ``Contact: Contact-aware tactile learning for robotic
  disassembly,'' \emph{arXiv preprint arXiv:2603.08560}, 2026.

\bibitem{wang2021gelsight}
S.~Wang, Y.~She, B.~Romero, and E.~Adelson, ``Gelsight wedge: Measuring
  high-resolution 3d contact geometry with a compact robot finger,'' in
  \emph{2021 IEEE International Conference on Robotics and Automation
  (ICRA)}.\hskip 1em plus 0.5em minus 0.4em\relax IEEE, 2021, pp. 6468--6475.

\bibitem{he2016deep}
K.~He, X.~Zhang, S.~Ren, and J.~Sun, ``Deep residual learning for image
  recognition,'' in \emph{Proceedings of the IEEE conference on computer vision
  and pattern recognition}, 2016, pp. 770--778.

\end{thebibliography}
}

\end{document}